\documentclass[11pt,a4paper]{article}
\usepackage[hyperref]{eacl2021}
\usepackage[utf8]{inputenc}
\usepackage[T2A,T1]{fontenc}
\usepackage[russian,english]{babel}
\usepackage{times}
\usepackage{latexsym}

\usepackage{graphicx}

\usepackage{microtype}

\aclfinalcopy 

\title{Representing ELMo embeddings as two-dimensional text online}

\author{Andrey Kutuzov \\
  University of Oslo \\
  \texttt{andreku@ifi.uio.no} \\\And
  Elizaveta Kuzmenko \\
  University of Trento \\
  \texttt{lizaku77@gmail.com} \\}

\date{}

\begin{document}
\maketitle
\begin{abstract}
We describe a new addition to the \textit{WebVectors} toolkit which is used to serve word embedding models over the Web. The new \textit{ELMoViz} module adds support for contextualized embedding architectures, in particular for ELMo models. The provided visualizations follow the metaphor of `two-dimensional text' by showing lexical substitutes: words which are most semantically similar in context to the words of the input sentence. The system allows the user to change the ELMo layers from which token embeddings are inferred. It also conveys corpus information about the query words and their lexical substitutes (namely their frequency tiers and parts of speech). The module is well integrated into the rest of the \textit{WebVectors} toolkit, providing lexical hyperlinks to word representations in static embedding models. Two web services have already implemented the new functionality with pre-trained ELMo models for Russian, Norwegian and English.
\end{abstract}





\section{Introduction} \label{sec:intro}
In this demo paper we describe a new module recently added to the free and open-source \textit{WebVectors} toolkit \citep{kutuzov2017building}\footnote{A screencast is available at \texttt{\url{https://www.youtube.com/watch?v=dDugoV1r\_wk}}.}. \textit{WebVectors} allows to easily deploy services to demonstrate the abilities of static distributional word representations (word embeddings) \citep{bengio2003neural,mikolov-etal-2013-linguistic} via web browsers. It currently powers at least two embedding model hubs:
\begin{itemize}
    \item \textit{NLPL WebVectors}\footnote{\url{http://vectors.nlpl.eu/explore/embeddings/}}, featuring models for English, Norwegian and other languages, trained within the Nordic Language Processing Laboratory initiative.
    \item  \textit{RusVect\={o}r\={e}s}\footnote{\url{https://rusvectores.org/}}, featuring models for the Russian language.
\end{itemize}

The new module (we name it \textit{ELMoViz}) adds the functionality to study, probe and compare recently introduced contextualized embedding (or `token-based') models \citep{melamud-etal-2016-context2vec}. In particular, at this point we provide support for the ELMo architecture \citep{peters-etal-2018-deep} based on deep recurrent neural networks. In the future, we plan to add support for Transformer-based models like BERT \citep{devlin-etal-2019-bert} and GPT-3 \citep{brown2020language}. ELMo architecture is significantly less computationally expensive than Transformers, while being almost on par in terms of performance. Thus, it yields rich possibilities in the context of non-commercial web services.

\begin{figure}
    \centering
    \includegraphics[width=\linewidth]{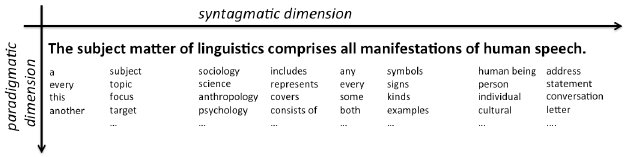}
    \caption{Metaphor of two-dimensional text; borrowed from \cite{biemann2013text}.}
    \label{fig:2d}
\end{figure}

For analyzing ELMo representations of an arbitrary input text, we offer the metaphor of `two-dimensional text' first proposed in \cite{biemann2013text} (see Figure~\ref{fig:2d}). This allows a sort of `visualization' for contextualized embeddings through finding words which are most semantically similar to the input words in their current contexts. From the linguistic point of view, these are `paradigmatic replacements' \citep{saussure1993course} -- words that can to some extent substitute target words. The two dimensions here are the \textit{syntagmatic} one (horizontal) which describes the linear order of the sentence, and the \textit{paradigmatic} one (vertical) which describes semantic classes to which the words in the sentence belong to. The generated substitutes in the vertical axis can also be thought of as `semantic variations' of the input sentence.

The rest of the paper is organized as follows. In Section~\ref{sec:related} we describe the background for this work, including the \textit{WebVectors} framework, and explain the need to develop additional functionality in order to handle contextualized embeddings. Section~\ref{sec:system} describes in detail this functionality, both from the point of view of the end user and from the point of view of deployment logistics. In Section~\ref{sec:conclusion}, we conclude and outline future work.

\section{Background} \label{sec:related}
Since the widespread adoption of prediction-based word embeddings \citep{mikolov-etal-2013-linguistic} started, there has always been a need to efficiently serve and demonstrate these representations over the Web. Researchers and practitioners need this for quick experimentation and testing hypotheses by comparing different distributional models. Those who teach natural language processing and computational linguistics need ways to show the students how dense distributional representations capture lexical semantics without installing any software or downloading any models (often it is desirable that this is shown for a particular language or domain).

In turn, language teachers value tools to demonstrate lexical variety and degrees of similarity for words in a foreign language. To this extent, serving word embeddings over the Web can help both the teachers with preparing educational materials and the students with grasping the concepts in a foreign language.

\begin{figure}
    \centering
    \includegraphics[width=\linewidth]{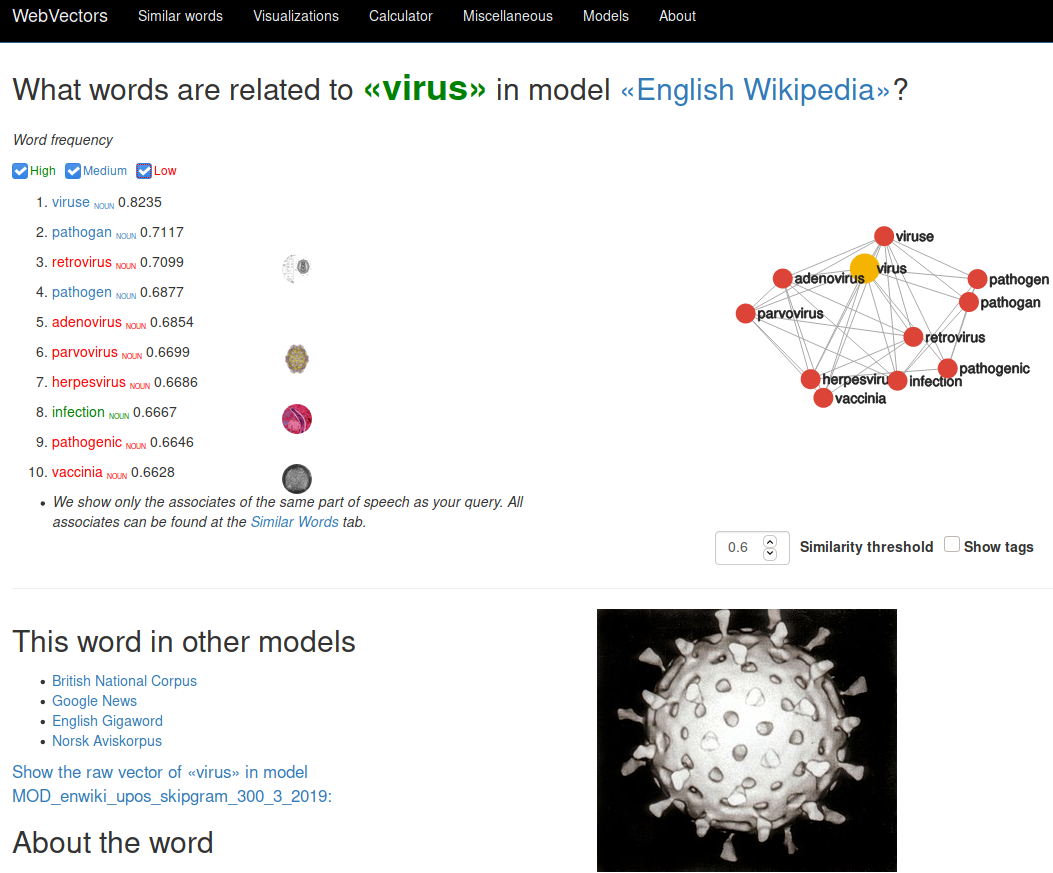}
    \caption{Screenshot of a \textit{WebVectors} instance at \url{http://vectors.nlpl.eu/explore/embeddings/}}
    \label{fig:wv}
\end{figure}

The \textit{WebVectors} framework we presented in \cite{kutuzov2017building} is aimed at all these purposes. It allows to quickly deploy a stable and robust web service featuring operations on vector semantic models, including querying, visualization and comparison, all available to users of any computer literacy level. It extended already existing embedding visualization services like Embedding Projector\footnote{\url{https://projector.tensorflow.org/}} by providing users with the ability to find nearest semantic neighbors of query words, perform vector math operations over embeddings, etc. Since being first presented in 2016, \textit{WebVectors} keeps adding new functionality, and now it offers filtering nearest associates by part of speech tags or corpus frequency, and can generate semantic ego graphs, among other features (see Figure~\ref{fig:wv}).

Until the introduction of \textit{ELMoViz}, these features were limited to the so-called `static word embeddings', that is, architectures like \textit{word2vec} \citep{mikolov-etal-2013-linguistic}, \textit{fastText} \citep{bojanowski-etal-2017-enriching} or \textit{GloVe} \citep{pennington-etal-2014-glove}. In these architectures, after the training is finished, each word type in the vocabulary is rigidly associated with a single dense vector. However, in the recent years NLP saw a surge of pre-trained `contextualized' embedding architectures, like ELMo \citep{peters-etal-2018-deep}, BERT \citep{devlin-etal-2019-bert}, GPT-3 \citep{brown2020language} and many others. One of the changes these deep learning models brought was that even at inference time, each word token representation (embedding) depends on its immediate context. This means that ambiguous words will receive different representations depending on the sense in which they are used, which opens rich new possibilities for natural language understanding.

Libraries used in \textit{WebVectors} to deal with static word embeddings (\textit{Gensim}, \citep{rehurek:2010}) were not fit to power operations on contextualized models. That is why we decided to implement an entirely new \textit{WebVectors} module, which would take a query phrase as an input, and produce paradigmatic replacements (lexical substitutions) for each content word in this phrase, based on a given pre-trained contextualized ELMo language model. 

One can find a number of existing frameworks for online experimentation with contextualized models: among others, we should mention Language Interpretability Tool \citep{tenney2020language}, exBert by \cite{hoover2019exbert} and the hosted inference API at the HuggingFace Community Model Hub \citep{wolf-etal-2020-transformers}. However, these projects are aimed exclusively at the Transformer-based architectures. The system we present in this demo paper, on the other hand, is aimed more towards RNN-based architectures like ELMo. As it was shown, for example, in the field of semantic change detection \citep{kutuzov-giulianelli-2020-uio}, ELMo can often outperform BERT or be on par with it, while requiring significantly less computational resources. We believe it is especially important for teaching activities.

Additionally, our system is more lexically oriented and is integrated with the existing \textit{WebVectors} functionality, as we will show in the next section.

\section{System description} \label{sec:system}
After turning on the contextualized embedding related functionality in the  \textit{WebVectors}  configuration file,\footnote{In principle, it is also possible to use only ELMoViz, without other \textit{WebVectors} modules.} the person deploying the service has to provide three data sources for each ELMo model:
\begin{enumerate}
    \item a pre-trained \textit{ELMo model} itself in the standard format (\texttt{*.HDF5} file with the weights and \texttt{options.json} file with the model architecture description);
    \item a tab-separated \textit{frequency dictionary} file to use when determining the frequency tier of word types (it is recommended to derive it from the same corpus the ELMo model was trained on, but technically this is not required);
    \item a set of \textit{static (type-based) word embeddings} produced by averaging contextualized token embeddings inferred with the same ELMo model.
\end{enumerate}

\begin{figure}
    \centering
    \includegraphics[width=\linewidth]{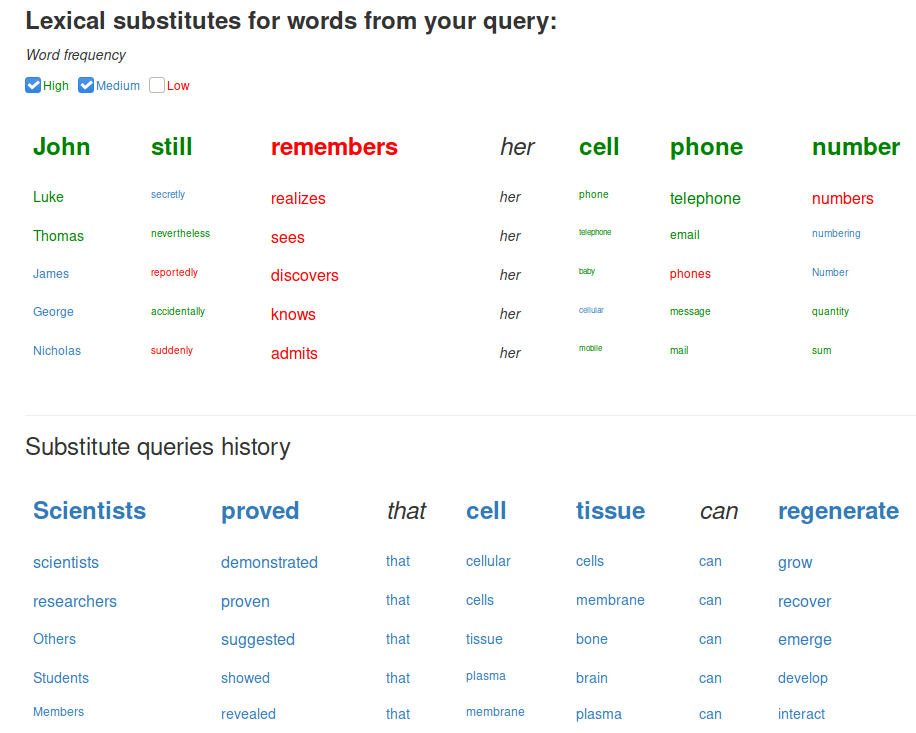}
    \caption{Examples of two-dimensional text inferred from an ELMo model ($n = 5$).}
    \label{fig:substitutes}
\end{figure}

The last item of this list requires some explanation. Our aim is to provide the end user with a set of lexical substitutes for each word token in context from the input sentence (see Figure~\ref{fig:substitutes}). With static embedding architectures, this boils down to looking up the vector of the target word $x$ and then finding $n$ other words in the model vocabulary with the vectors closest to $x$. However, this is obviously impossible with contextualized language models: there are no static vector lookup tables to begin with. One can easily infer contextualized representations for each word in the input sentence: but what to compare them with in order to illustrate their meaning? 

To cope with this issue, we adopted the approach described in \citep{liu-etal-2019-investigating}. They employed the so called type-level context averaging in order to  align pre-trained contextualized models cross-linguistically. In our case, we needed only the first stage of their workflow. The idea is to obtain static type-level word representations located in the same vector space as the contextualized embeddings. Given a large enough reference text corpus and a pre-trained contextualized language model, one takes the average of all token representations for each target word occurrence in the corpus. This averaged type embedding is comparable to contextualized token embeddings routinely produced by the model.

In practice, we found that one does not even need to \textit{average} token embeddings: it is enough to \textit{sum} them, and then unit-normalize the resulting summed vector. As for the list of target words, we simply use top 10 000 (or any other amount found suitable) most frequent words from the corresponding ELMo model vocabulary or from a reference corpus (excluding functional parts of speech and digits). Low frequent words are usually not needed in this case anyway, since the quality of their embeddings is also lower. We provide a simple script to extract type embeddings from an ELMo model and a given corpus in our GitHub repository.\footnote{\url{https://github.com/akutuzov/webvectors/tree/master/elmo/}}

As a result, when an end user enters an input phrase or sentence (typically from 5 to 15 words), \textit{WebVectors} produces contextualized token embeddings for each token in the query, and finds top $n$ words in the \textit{type embedding} model, which are the closest (by cosine similarity) to each of the \textit{token embeddings}. These predictions are \textit{lexical substitutes} or \textit{paradigmatic replacements}; they demonstrate what other words could fill these positions in the query, depending on the context.

Another option to produce such substitutes would be to feed the input sentence to the ELMo model and then for each word token choose the strongest activations at the final softmax layer of the language model and map them to words in the model vocabulary. However, in practice we found that this approach is slightly slower than the one described above. Additionally, ELMo models are often published online without the vocabulary they were trained on. Since the input layer of ELMo is purely character-based, it does not hinder inferring token embeddings, but it effectively blocks using these weights as language models \textit{per se}. Our approach allows one to use any given ELMo model with any desired corpus to produce a set of reference type embeddings.

System maintainers can provide several models for the service to work with, including models for different languages; one of the models should be specified in the configuration files as the default one. When entering the query sentence, users can choose the model which will process the input. 

Apart from choosing between different models, \textit{WebVectors} also allows users to choose the exact ELMo layer from which token representations will be inferred; it was shown in \cite{peters-etal-2018-dissecting}  that different neural network layers convey information related to different linguistic tiers: syntax, semantics, pragmatics, etc. At this point, one can choose between the top ELMo layer and the average of all layers. Note that for all operations with pre-trained ELMo models we use \texttt{simple\_elmo}: a lightweight TensorFlow-based Python package also developed by us.\footnote{\url{https://pypi.org/project/simple-elmo/}} If need be, \texttt{simple\_elmo} can also be used as a standalone library to handle ELMo models.

Both the words from the input sentence and the lexical substitutes are colored according to their frequency tier in the reference corpus (\textcolor{green}{green} for `high', \textcolor{blue}{blue} for `mid' and \textcolor{red}{red} for `low'), in accordance with other \textit{WebVectors} components. Similarly, each word is hyperlinked to its `landing page' bound to one of the static embedding models served by a particular \textit{WebVectors} installation (like the one in Figure~\ref{fig:wv}), allowing easy and playful exploration of the semantic space. The font size of the lexical substitute corresponds to cosine similarity between the token embedding and the substitute type embedding: thus, users can instantly see what word tokens the model is unsure about. The service performs fast under-the-hood part-of-speech tagging of the query,\footnote{Using UDPipe \citep{straka-strakova-2017-tokenizing}.} so for functional words we always yield themselves as substitutes (see `her', `that' and `can' in Figure~\ref{fig:substitutes}). They are also uncolored and not hyperlinked, so that a user might focus on content words, while at the same time still having an impression of `full sentence variations'.

The users should be aware that the lexical substitutes potentially contain all the biases inherited from the corpus the model was trained on. Thus, the paradigmatic axis might include slander words and stereotypes, if they were frequent enough in the data. We did not address this issue in the present work, but we advise the users to take this into account when dealing with any unsupervised language models.

\begin{figure}
    \centering
    \includegraphics[width=\linewidth]{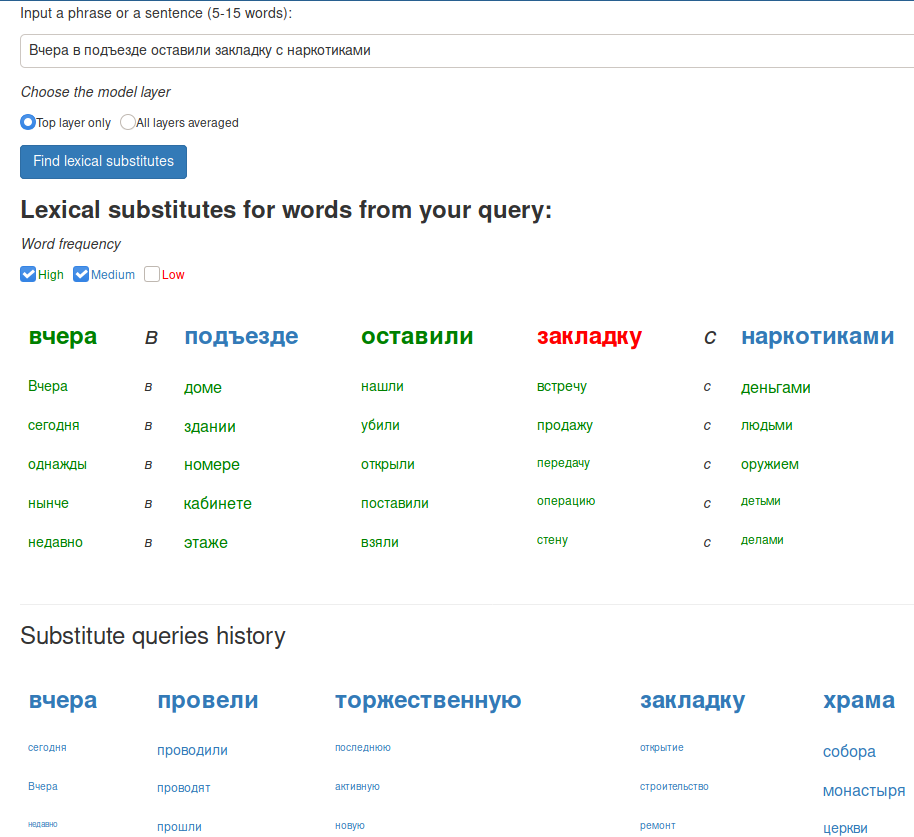}
    \caption{History of lexical substitute queries with a Russian ELMo model.}
    \label{fig:zaklad}
\end{figure}

Importantly, we keep a short history of substitute queries, so that it is possible to see at a glance the changes brought by a different context, a different word order or a different contextualized model (if the web service offers several models). Figure~\ref{fig:zaklad} shows an example from our Russian live demo at the \textit{RusVectōrēs} web service. In the first sentence, the word \foreignlanguage{russian}{закладку} `\textit{zakladku}' is used in the newer sense of `a secret place to store illegal drugs', while in the second sentence it is used in the older sense of `the act of founding a building'. The generated substitutes reflect the differences in word meaning depending on the context. In the first example the substitutes include such words as `meeting, sale, operation', and in the second example the substitutes are `opening, building, repair'.

\section{Conclusion} \label{sec:conclusion} 
The described system for generating two-dimensional text using pre-trained ELMo models is now deployed at the two model hubs mentioned in Section~\ref{sec:intro}. \textit{NLPL WebVectors} features ELMo models trained on English Wikipedia and on Norwegian corpora\footnote{\url{http://vectors.nlpl.eu/explore/embeddings/en/contextual}}, while \textit{RusVect\={o}r\={e}s} features a model  trained on concatenated Russian Wikipedia and Russian National Corpus.\footnote{\url{https://rusvectores.org/en/contextual/}} 

The presented component for the \textit{WebVectors} framework allows users to explore pre-trained ELMo models and to visualize contextualized embeddings as a two-dimensional text for faster analysis of early research prototypes. While previously the framework provided interface only to static vector semantic models, introducing support for contextualized architectures allows for more intricate exploration of linguistic phenomena, such as lexical ambiguity and contextual semantic change.

We hope that the new functionality will provide language teachers, NLP researchers and practitioners with a powerful tool to study word meaning in context and at the same time keep the audience up-to-date with recent advances in the field of distributional semantics and deep learning based NLP. A separate important contribution is our \texttt{simple\_elmo} library which makes using ELMo models in Python much easier, especially for researchers with linguistic background.

In the future, we plan to add support for other contextualized embedding architectures like BERT, to allow inter-architectural comparisons. 
Another interesting room for future work is integrating with other exploratory services for neural NLP models, like the ones mentioned in Section~\ref{sec:related}.

\bibliography{anthology,eacl2021}
\bibliographystyle{acl_natbib}

\appendix

\end{document}